# Global Preferential Consistency for the Topological Sorting-Based Maximal Spanning Tree Problem[a]

Rémy-Robert Joseph[*]

**Abstract.** We introduce a new type of fully computable problems, for DSS dedicated to maximal spanning tree problems, based on deduction and choice: *preferential consistency* problems. To show its interest, we describe a new compact representation of preferences specific to spanning trees, identifying an efficient maximal spanning tree sub-problem. Next, we compare this problem with the Pareto-based multiobjective one. And at last, we propose an efficient algorithm solving the associated preferential consistency problem.

## 1 INTRODUCTION

Given an undirected graph $G = (V, E)$ with $V$ the vertices and $E$ the edges, a **spanning tree** $x$ of $G$ is a connected and acyclic partial graph of $G$. $x$ is then always composed with $|V| - 1$ edges. We denote by $S_{ST}(G)$ the spanning trees set of $G$. For short, we write: $e \in x$, with $e \in E$, to say: $e$ is an edge of the spanning tree $x$. More generally, we will assimilate $x$ to its edges set. The classical problem of maximum spanning tree ($\Leftrightarrow$ ST/$\Sigma u$/OPT) is defined as follow:

ST/$\Sigma u$/OPT: Given an undirected graph $G = (V, E)$ and a utility $u(e)$ associated with each edge $e \in E$, the result is a feasible spanning tree $x$ of $G$, maximizing the sum of utilities of edges in $x$, if such a tree exists. Otherwise, the result is 'no'.

Several consistency problems have been recently investigated on spanning trees. On the one hand, we note the consistency problem associated with feasible spanning trees of a graph [25]. Other investigations pointed out consistency associated with weighted spanning trees [8], and maximum spanning tree [9]. On the other hand, numerous local consistency problems combining classical spanning tree problems with other constraints have been investigated. For example, the diameter constrained minimum spanning tree problem (DCMST) [16].

Within non-conventional preferences, the situation is radically different. Very few consistency spanning tree problems have been investigated in literature. We cite a local consistency problem processed for the robust spanning tree problem with interval data (RSTID) [1].

Yet, the most of combinatorial problems from the real practical world require the modeling of imprecision or uncertainty, multiple divergent viewpoints and conflicts management, to wholly assess the solutions and to identify the best compromise ones. These singularities require more complex modeling of preferences [27, 21]. For now some decades, the OR/CP community scrutinizes combinatorial problems enabling non-conventional global preferences. Thus, we attended to the flowering of a great number of publications dealing with multiobjective combinatorial optimization problems (see [10, 3] for surveys). Nevertheless, a very few articles dealt with combinatorial problems with purely ordinal and/or intransitive preferential information. We mention the recent investigations in the scope of (i) decision theory with maximal spanning trees and maximal paths in a digraph [18], (ii) game theory with stable matchings (see [20] for a survey), (iii) algebraic combinatorial optimization [28, 5], or (iv) artificial intelligence with some configuration problems [4, 14] and with heuristic search algorithms [17, 14]. We decide to bring another stone to this building, with the concept of preferential consistency applied to the topological sorting-based maximal spanning trees problem.

The decision problematic of finding a suitable preferred solution is semi-structured: in the general case (beyond total preorders), a preferred solution fitted to the decision-maker cannot be only identified from the implemented preferential information. Preferred solutions are not all equivalent, some are partially comparable others are incomparable, and sometimes, there exists no optimal or maximal solution [27, 21]. To investigate these semi-computable problems, we will use the concept of Decision Support System (DSS) to explore the preferred solutions set. This exploration can be achieved other than by building iteratively new preferred solutions – as usually in multicriteria optimization –; For example, by describing this preferred set with the set of values present in at least one preferred solution. The notion of consistency, defined in Constraint Programming, gathers the theoretical surrounding of this descriptive approach of implicit sets. This is a reactive [26] and deductive approach of solving; In a polynomial number of actions (removings, instantiations and backtrackings), the user leads to a preferred solution.

Consequently, after an introduction on preference relations (§ 2.1), we make a brief presentation on compact representation of preferences (§ 2.2). We next point out a generalization of the maximum spanning trees problem: the maximal spanning trees problem (§ 3.1). So, we introduce (§ 3.2) preferential consistency,

---

i.e. a template redefining consistency in order to take into account of peculiarities of combinatorial problems exploiting non-conventional preferences, followed by its using on the maximal spanning trees. In general, most of relevant computable problems supporting the initial decision problem are intractable. Accordingly, we point out an <u>easy</u> suitable maximal spanning trees sub-problem (§ 4), based on a compact preference representation inspired by topological sorting (§ 4.1). In § 4.2, we give an example of using in the multicriteria context and we compare this sub-problem with the Pareto-based multiobjective version. Next, we design a global preferential consistency algorithm (§ 5) dedicated to it. We conclude (§ 6) with some perspectives.

## 2 PREREQUISITES IN DECISION THEORY

Throughout this article, we take place at a very general abstraction level, where global preferences are represented by a non complete, intransitive and even cyclic binary relation on the solutions space, but enabling a <u>maximal set</u> (there exist no solution strictly preferred to any of them). Here are some definitions:

### 2.1 Preference relation

Given a non-empty finite set $S$, a **(crisp binary) preference relation** [23, 27, 21] $\succcurlyeq$ of an individual on $S$ is a reflexive binary relation on $S$ ($\Leftrightarrow \succcurlyeq \subseteq S \times S$ and $\forall x \in S, (x, x) \in \succcurlyeq$) translating some judgments of this individual concerning his preferences between the alternative elements of $S$. For every couple of elements $x$ and $y$ of $S$, the assertion « $x \succcurlyeq y$ » is equivalent to « $(x, y) \in \succcurlyeq$ » and means that « $x$ is at least as good quality as $y$ for considered individual ». A preference relation $\succcurlyeq$ carries out a partition of $S \times S$ into four fundamental relations:

(**indifference**) $x \simeq y \Leftrightarrow (x \succcurlyeq y$ and $y \succcurlyeq x)$ for all $x, y \in S$
(**strict preference**) $x \succ y \Leftrightarrow (x \succcurlyeq y$ and $\text{not}(y \succcurlyeq x))$ for all $x, y \in S$
(**strict aversion**) $x \prec y \Leftrightarrow y \succ x$ for every $x, y \in S$
(**incomparability**) $x \parallel y \Leftrightarrow (\text{not}(x \succcurlyeq y)$ and $\text{not}(y \succcurlyeq x))$ for every $x, y \in S$

Preference relations defined on a finite set formally correspond with the concept of simple directed graphs (shortly digraphs). Accordingly, the graphical representation of digraphs will allow us to illustrate our investigation. For every non-empty $A \subseteq S$, the **restriction** of $\succcurlyeq$ to $A$ is the preference relation $\succcurlyeq_{|A}$ defined as follow: $\succcurlyeq_{|A} = \{(x, y) \in A \times A,$ such that: $x \succcurlyeq y\}$. By abuse, we do not specify the restriction, the context enabling to identify the targeted subset of $S$. A preference relation $\succcurlyeq$ is:

- **transitive** iff $[x \succcurlyeq y$ and $y \succcurlyeq z] \Rightarrow x \succcurlyeq z$, for all $x, y, z \in S$
- **quasi-transitive** iff $[x \succ y$ and $y \succ z] \Rightarrow x \succ z$, for all $x, y, z \in S$ iff the strict preference relation is transitive
- **P-acyclic** iff $\forall t > 2$ and $\forall x_1, x_2, ..., x_t \in S$,
  $$[x_1 \succ x_2 \succ ... \succ x_t] \Rightarrow \text{not}(x_t \succ x_1)$$
  iff $\succcurlyeq$ has no circuit of strict preference.
- an **equivalence relation** iff it is reflexive[1], symmetric and transitive
- a **partial preorder** iff it is reflexive and transitive

- a **complete** (or **total**) **preorder** iff it is reflexive, transitive and complete
- a **complete** (or **total**) **order** iff it is reflexive, transitive, antisymmetric and complete

Given a finite non-empty set $S$ structured by a preference relation $\succcurlyeq$, the **maximal set** (or **efficient set**) of $S$ according to $\succcurlyeq$, denoted $M(S, \succcurlyeq)$, is the subset of $S$ verifying: $M(S, \succcurlyeq) = \{x \in S \mid \forall y \in S, \text{not}(y \succ x)\}$; while the **optimal set** of $S$ according to $\succcurlyeq$, denoted $B(S, \succcurlyeq)$, is the subset of $S$ verifying: $B(S, \succcurlyeq) = \{x \in S \mid \forall y \in S, x \succcurlyeq y\}$. Of course, there exists other choices of axioms identifying preferred (i.e. best quality, or best compromise) solutions from a preference relation, and we refer to [11, 24] for a deepening.

Given a preference relation $\succcurlyeq$ on a finite set $S$, another preference relation $\succcurlyeq'$ on $S$ is an **extension** of $\succcurlyeq$ if $\forall x, y \in S, x \succ y \Rightarrow x \succ' y$. The relation $\succcurlyeq'$ is called a **linear extension** of $\succcurlyeq$ if $\succcurlyeq'$ is an extension of $\succcurlyeq$ and $\succcurlyeq'$ is a total order. We have the following result (see [23]): a preference relation $\succcurlyeq$ on a finite set $S$ is P-acyclic $\Leftrightarrow$ every non empty subset of $S$ has a non empty maximal set ($\Leftrightarrow \forall \emptyset \neq A \subseteq S, M(A, \succcurlyeq) \neq \emptyset$) $\Leftrightarrow$ there exists linear extensions of $\succcurlyeq$ and they are obtained by topological sorting.

### 2.2 Compact representations of preferences in combinatorial problems

In combinatorial practical applications, solutions are implicit: described by a set $S$ of elementary components of a set $E$ ($\Leftrightarrow S \subseteq \mathcal{P}(E)$). Then, it is necessary to imagine a **compact representation of preferences** for their elicitation (acquisition) and their processing; because these operations with an explicit representation – the listing of the couples $x, y \in S$ such that $x \succcurlyeq y$ – being usually intractable.

Thus, in classical combinatorial optimization, the preferences are represented by a utility function $u$ from $\mathcal{P}(E)$ to $\mathbb{R}$ to maximize: $x \succcurlyeq y \Leftrightarrow u(x) \geq u(y)$. In multicriteria optimization based on the Pareto dominance, preferences are represented by a vector of utility functions $(u_1, ..., u_p)$, aggregated by the Pareto dominance: $x \succcurlyeq y \Leftrightarrow [\forall i \in \{1, ..., p\}, u_i(x) \geq u_i(y)]$. This hierarchical aggregation will be noted $p\Sigma u_{\triangleright \text{PARETO}}$. And more generally, every aggregation of a family of $p$ utility functions by a rule $AR$ will be noted $p\Sigma u_{\triangleright AR}$. In artificial intelligence, numerous compact representations of preferences appeared: from CP-nets [4, 14] to constraints describing the preferential neighbourhood of the solutions (called preferential constraints in [13]), by going through soft constraints [3, 19] and dynamic CSP [26]. In the following, any compact representation of a preference relation $\succcurlyeq$ is denoted $I(\succcurlyeq)$. We will present in § 4.1 the compact representation used here for our maximal spanning trees sub-problem.

## 3 PREFERENTIAL CONSISTENCY AND MAXIMAL SPANNING TREES

### 3.1 Maximal spanning trees problems

Consider the problem of finding a *satisficing* (in the meaning of Newell & Simon [15]) maximal spanning tree. Denoted by

---

[1] Mention that a binary relation $\succcurlyeq$ is **symmetric** iff $x \succcurlyeq y \Rightarrow y \succcurlyeq x$, for all $x, y \in S$; **antisymmetric** iff $x \succcurlyeq y \Rightarrow \text{not}(y \succcurlyeq x)$, for all $x, y \in S$ with $x \neq y$; and **complete** (or **total**) iff $x \succcurlyeq y$ or $y \succcurlyeq x$, for all $x, y \in S$ and $x \neq y$.

DS(ST/CBPR/MAX), this semi-structured problem is formulated in the following way:

DS(ST/CBPR/MAX): Given an undirected graph $G = (V, E)$ and a compact representation $I(\succcurlyeq)$ of a preference relation $\succcurlyeq$ on $\mathcal{P}(E)$, the result is a feasible spanning tree which is:
  (i) maximal for $(S_{ST}(G), \succcurlyeq)$, if such a solution exists, and
  (ii) suited with the system of values of the user.
Otherwise, the result is 'no'.

*Remark 1.* DS and CBPR mean respectively decision support and crisp binary preference relation. The condition (ii) means the user via an interactive process will treat the lack of equivalence and the incompleteness between maximal solutions. This definition of problem involves that the *satisficing* solution, must be also maximal in $(S_{ST}(G), \succcurlyeq)$. In other words, the only degree of freedom let to the DSS user is the choice of a suited solution among the maximal ones. This definition refers for example to contexts where preferences have been given by the different actors of the decision problem, next aggregated in global – possibly incomplete and intransitive – preferences $\succcurlyeq$ on the solutions $\mathcal{P}(E)$ via a compact representation $I(\succcurlyeq)$; Now, an individual: the user, being able to bring efficiently forgotten preferential information at different times of the decision process, is in charge of finding the suited solution mirroring at best global preferences.

At this semi-structured problem is associated the computable problem of finding a maximal spanning tree, denoted ST/CBPR/MAX, the definition of which corresponds with the DS(ST/CBPR/MAX) one, after erasing the property (ii). In such a general framework, these computable problems are hard. To be convinced, it is sufficient to consider the peculiar case where the used compact representation of preferences is the Pareto-based multicriteria one. Hence, the membership problem associated with this multiobjective spanning trees problem is NP-complete [6, 12].

## 3.2 Preferential consistency for maximal spanning trees

In Constraint Programming [19], **consistency** is a part of a more general problematic called **description**. The aim of consistency is the description of the feasible set of a *constraint system* by way of values or combinations of values belonging to at least one feasible element.

Consistency problematic can be extended, in the framework of combinatorial problems exploiting non-conventional preferences, so as to take into account of preferential information. Simply, consistency will not rely on feasibility but on best quality or best compromise. Hence, we won't remove inconsistent values in the meaning that they belong to no feasible solution, but rather because they belong to no preferred solution. In this case, we speak about **preferential consistency**.

Without going into details, problems consisting in erasing preferentially inconsistent values, from a constraint system and a compact representation of a preference relation, are called **preferential consistency problems**. As in constraint satisfaction, several levels of preferential consistency can be defined, according to whether all or a part of preferentially inconsistent information is deleted. We named **global preferential consistency** the removing of all the preferentially inconsistent information.

*Remark 2.* In a non-conventional preference context, each used choice axiom (e.g. optimality, maximality, domination, …) identifies a specific choice set (optimal set, maximal set, domination set, …) which are generally pairwise different (see § 2.1). This other parameter specializes preferential consistency. Thus, we speak about OPT-consistency for preferential consistency using optimality as choice axiom, MAX-consistency for preferential consistency using maximality, and so on.

To better understand preferential consistency, in the following, we study in details the case of maximal spanning tree problem. Consider then the following general computable problem, of preferential consistency for maximal spanning trees of a graph:

GPC(ST/CBPR/MAX): Given an undirected graph $G = (V, E)$ and a compact representation $I(\succcurlyeq)$ of a preference relation $\succcurlyeq$ on $\mathcal{P}(E)$, list the edges in $E$ belonging to a maximal spanning tree for $\succcurlyeq$, if such edges exist. Otherwise return 'no'.

An edge $e$ is called MAX-consistent for $(G, I(\succcurlyeq))$ if there exists at least one maximal spanning tree for $(S_{ST}(G), \succcurlyeq)$ containing $e$. Otherwise, it is called MAX-inconsistent for $(G, I(\succcurlyeq))$.

In this article, we do not dwell on the computational complexity of this problem. But there are great chances it is at least as difficult as ST/CBPR/MAX, with the sight of investigations in constraint programming [2, 19]. Yet, in order to better appreciate the using of this kind of computable problem in a DSS, we turn towards an efficiently solvable sub-problem of ST/CBPR/MAX.

## 4 THE ST/TOSORT-VSMAX/MAX PROBLEM

### 4.1 Compact representation and TOSORT-VSMAX condition

From now, to point out an edges set, for example $\{a, b\}$, we adopt the notation $ab$. Given an undirected graph $G = (V, E)$ and a P-acyclic preference relation $\succcurlyeq_E$ on $E$, we consider the binary relation $\succcurlyeq_K$ on $\mathcal{P}(E)$ defined as follow:
$\forall x, y \in \mathcal{P}(E), x \succcurlyeq_K y \Leftrightarrow$
  $\exists$ a linear extension $\{e_1, …, e_{|E|}\}$ of $\succcurlyeq_E$ on $E$, verifying:
    $e_i \succ_E e_j \Rightarrow i < j$ for all $1 \leq i, j \leq |E|$, and
    for every $1 \leq j \leq |E|$, $e_j \notin x \Rightarrow (x \cap \{e_1, …, e_{j-1}\}) \cup \{e_j\}$ contains a cycle

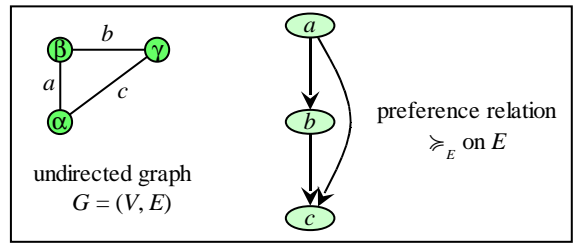

**Figure 1.** Example of an undirected graph and a totally ordered preference relation on its edge set[2].

**Example 1.** The Figure 1 illustrates the case of a complete undirected graph $G$ on 3 vertices, with a total order $\succcurlyeq_E$ on $E$. Then, the binary relation $\succcurlyeq_K$ verifies (Figure 2), in addition with reflexive arcs, that: $A \succ_K B$, for every $A \in \mathcal{M} = \{ab, abc\}$ and $B \in \mathcal{P}(E) \setminus \mathcal{M}$, because the only linear extension of $\succcurlyeq_E$ is itself.

---
[2] To avoid surcharges of the graphical representation, the reflexive arcs are not drawn.

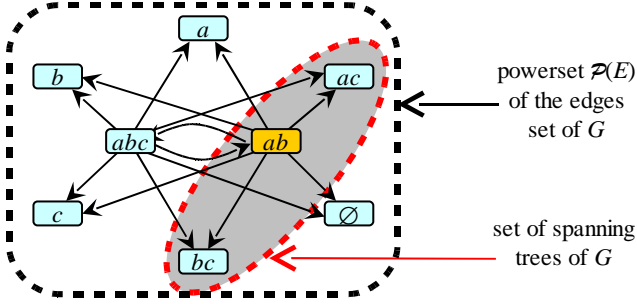

**Figure 2.** The relation $\succcurlyeq_K$ elaborated from $(G, \succcurlyeq_E)$ of Figure 1.

The Figure 3 considers an undirected graph $G = (V, E)$, with $V = \{\alpha, \beta, \gamma, \delta\}$ and $E = \{a, b, c, d, h\}$; and a P-acyclic relation $\succcurlyeq_E$ on the edges set $E$ of $G$ verifying, in addition of reflexive arcs: $a \succ_E h$, $c \succ_E b$, $c \succ_E d$, $d \succ_E b$, $h \succ_E b$, $c \simeq_E h$, $d \simeq_E h$.

Then, the binary relation $\succcurlyeq_K$ establishes a bipartition $\{\mathcal{M}, \mathcal{P}(E) \setminus \mathcal{M}\}$ of $\mathcal{P}(E)$ with $\mathcal{M} = \{x \in \mathcal{P}(E) \text{ such that: } acd \subseteq x \text{ or } ach \subseteq x\}$ and satisfies the following relations: $\forall (A, B) \in \mathcal{M} \times (\mathcal{P}(E) \setminus \mathcal{M})$, $A \succ_K B$ and $\forall (A_1, A_2) \in \mathcal{M} \times \mathcal{M}, A_1 \simeq_K A_2$.

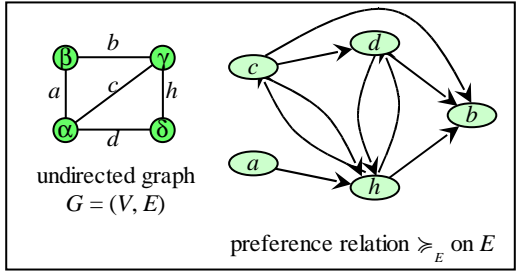

**Figure 3.** Example of an undirected graph and a P-acyclic preference relation on its edge set[32].

**Definition 1.** A preference relation $\succcurlyeq$ on $\mathcal{P}(E)$ is called TOSORT-VSMAX for the couple $(G, \succcurlyeq_E)$ iff: $\forall (x, y) \in S_{ST}(G) \times S_{ST}(G)$ with $x \neq y$,
$$\begin{cases} x \succ_K y \Rightarrow x \succ y & (\Leftrightarrow \text{ the relation } \succcurlyeq \text{ is an extension of } \succcurlyeq_K) \\ x \simeq_K y \Rightarrow x \simeq y \text{ or } x \parallel y \end{cases}$$

*Remark 3.* The word TOSORT in the notation TOSORT-VSMAX points out the relation $\succcurlyeq_K$: the relation $\succcurlyeq_E$ can be topologically sorted $\Leftrightarrow$ the relation $\succcurlyeq_E$ is P-acyclic $\Leftrightarrow$ there exists a non-empty maximal set of edges for every non-empty edges subset of $E$ $\Leftrightarrow$ there exist total orders extending $\succcurlyeq_E$. And the second word VSMAX points out both conditions of this definition – the extension condition and the translation of the indifference of $\succcurlyeq_K$ into indifference and incomparability of $\succcurlyeq$ – which define a very strong version of maximality.

**Example 2.** The Figure 4 illustrates a preference relation on $\mathcal{P}(E)$ satisfying TOSORT-VSMAX for the couple $(G, \succcurlyeq_E)$ of Figure 1. This illustration shows a TOSORT-VSMAX relation may include strict preference circuits.

For the Figure 3, the feasible spanning trees set is $S_{ST}(G) = \{abd, abh, acd, ach, adh, bcd, bch, bdh\}$; Accordingly, every TOSORT-VSMAX preference relation $\succcurlyeq$ on $\mathcal{P}(E)$ satisfies:
$$\begin{cases} \forall (x, y) \in \{acd, ach\} \times S_{ST}(G) \setminus \mathcal{M}, \; x \succ y \\ acd \text{ and } ach \text{ are either indifferent or incomparable} \end{cases}.$$

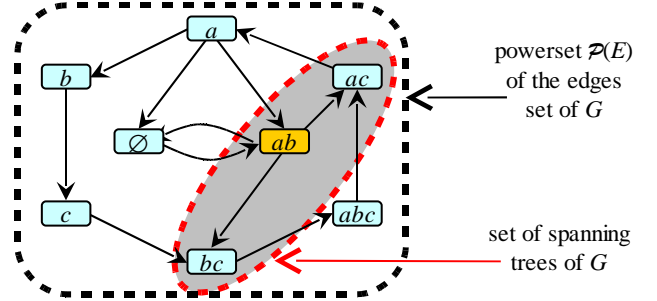

**Figure 4.** An example of TOSORT-VSMAX preference relation[2] on the powerset of $E$ of Figure 1.

The preference relation $\succcurlyeq_E$ on $E$ is called the *compact representation* of the TOSORT-VSMAX relation $\succcurlyeq$ on $\mathcal{P}(E)$. Here are some properties:

**Properties 1.** Given a couple $(G, \succcurlyeq_E)$ made up an undirected graph $G = (V, E)$ and a P-acyclic relation $\succcurlyeq_E$ on the edges set $E$, then:
(a) Every TOSORT-VSMAX preference relation for $(G, \succcurlyeq_E)$ identifies the same maximal set as the relation $\succcurlyeq_K$ induced by $\succcurlyeq_E$.
(b) The existence of feasible spanning trees warranties the existence of a non empty maximal set for $(S_{ST}(G), \succcurlyeq_K)$.

The proof is immediate. The relation $\succcurlyeq_K$ is the minimum information to know in order to identify the maximal set of TOSORT-VSMAX preference relations. Now, we consider the following sub-problem of ST/CBPR/MAX:

ST/TOSORT-VSMAX/MAX: Given an undirected graph $G = (V, E)$ and a compact representation $\succcurlyeq_E$ of a TOSORT-VSMAX preference relation $\succcurlyeq$ on $\mathcal{P}(E)$, return a maximal spanning tree for $\succcurlyeq$, if such a solution exists. Otherwise return 'no'.

We denote $S_{ST/TV/MAX}(G, \succcurlyeq_E)$ the set of possible maximal spanning trees outputted by an algorithm solving this problem.

**Theorem 1.** *The ST/TOSORT-VSMAX/MAX problem can be solved in a polynomial time in the input size $(G, \succcurlyeq_E)$.*

**Sketch of Proof:** One algorithm consists in elaborating a linear extension $\{e_1, \ldots, e_{|E|}\}$ of $\succcurlyeq_E$ on $E$ ($\Leftrightarrow$ the TOPOLOGICAL SORT problem[3] [7, 22]); Next in assigning a utility $u(e)$ to each edge $e$ of $E$ in order to satisfy the following condition: $u(e_i) > u(e_{i+1})$, $1 \leq i \leq |E| - 1$; for example, $u(e_i) = |E| - i$. And, at last in solving the classic spanning tree problem ($\Leftrightarrow$ ST/$\Sigma u$/OPT) with the instance $(G, u)$. The resulting

---

[3] In the rest of this article, we will have to use a particular algorithm solving this problem. We will consider the following one: increasingly and greedily number the maximal edges among the not yet numbered edges of $E$. The designed list of edges is then a linear extension of $\succcurlyeq_E$.

maximum spanning tree is then also a maximal solution for ST/TOSORT-VSMAX/MAX. ❑

## 4.2 Multiobjective spanning tree problems based on topological sorting

Now we confront this problem to the classical maximum spanning tree problem, and its Pareto-based multiobjective version.

**Example 3.** The classical problem of maximum spanning tree (⇔ ST/Σ$u$/OPT) can be polynomially transformed into the ST/TOSORT-VSMAX/MAX problem. Indeed, for any spanning tree $x$ of $G$, the sum of utilities of edges in $x$ defines a total preorder $\succcurlyeq_u$ on $\mathcal{P}(E)$:

$$\forall (x, y) \in \mathcal{P}(E)^2, x \succcurlyeq_u y \Leftrightarrow \sum_{e \in x} u(e) \geq \sum_{e \in y} u(e)$$

The relation $\succcurlyeq_u$ is TOSORT-VSMAX, and its compact representation $\succcurlyeq^u_E$ is the preorder induced by $u$: $\forall e, e' \in E, e \succcurlyeq^u_E e' \Leftrightarrow u(e) \geq u(e')$. The couple $(G, \succcurlyeq^u_E)$ is then an instance of ST/TOSORT-VSMAX/MAX, and its solution set $S_{ST/TV/MAX}(G, \succcurlyeq^u_E) = B(S_{ST}(G), \succcurlyeq_u)$. This assertion is easily provable by erasing the topological sorting part of the sketch of proof of Theorem 1.

The ST/TOSORT-VSMAX/MAX problem can be used to model and solve multicriteria problems. So, the multi-attribute utility function can be aggregated first to produce global preferences on the edges, and next to partially rank sets of edges. Here is an example:

**Example 4.** The ST/PARETO>TOSORT-VSMAX/MAX problem considers an undirected graph $G = (V, E)$ and a couple $(p, u)$ made up a positive number $p$ and a multi-attribute utility function $u$ from $E \times \{1, ..., p\}$ to ℝ. $p$ is the number of considered criteria and $u(e, k)$ is the utility of the edge $e$ according to the criterion $k$. In this problem, the preference information $(p, u)$ is aggregated with Pareto dominance, in order to define a global preference relation $\succcurlyeq_{EP}$ on each edge:

$\forall e, e' \in E, e \succcurlyeq_{EP} e' \Leftrightarrow$ for every $1 \leq k \leq p, u(e, k) \geq u(e', k)$

Next, this preference relation on the edges is aggregated with the $\succcurlyeq_K$ relation, to obtain a collective opinion $\succcurlyeq_{PK}$ between the subsets of $E$.

Then we consider the instance $(G, (2, u))$ made up the undirected graph $G = (V, E)$ of the Figure 3, and the bicriteria utility function $u$ given by the following table:

**Table 1.** Example of bicriteria utility function $u$(edge, criterion) on the edges of the undirected graph of the Figure 3.

|  | edges |  |  |  |  |
|---|---|---|---|---|---|
|  | $a$ | $b$ | $c$ | $d$ | $h$ |
| criterion 1 | 2 | 2 | 1 | 1 | 3 |
| criterion 2 | 1 | 1 | 3 | 2 | 0 |

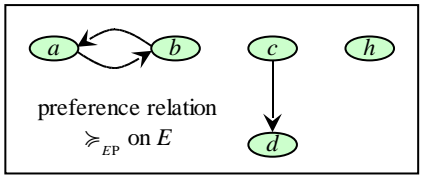

preference relation $\succcurlyeq_{EP}$ on $E$

**Figure 5.** The preference relation $\succcurlyeq_{EP}$ on $E$ provided by aggregation of $u$ with the Pareto dominance[32].

By aggregating $u$ with the Pareto dominance, we obtain the preference relation $\succcurlyeq_{EP}$ on $E$ given by the Figure 5. At last, by solving the ST/TOSORT-VSMAX/MAX problem on this instance $(G, \succcurlyeq_{EP})$, we get the maximal set $M(S_{ST}(G), \succcurlyeq_{PK}) = \{abd, abh, acd, ach, bcd, bch\}$

*Remark 4.* Instead of using the Pareto dominance to obtain the global preference relation $\succcurlyeq_{EP}$ on the edges, we can apply any aggregation rule $AR$ on $u$. The only condition on $AR$ is to provide a preference relation $\succcurlyeq_{EP}$ having at least the P-acyclicity property.

In the multicriteria decision-making community [10], the multi-attribute utility function $u(e, k)$, with $(e, k) \in E \times \{1, ..., p\}$, is usually aggregated with a simple sum per criterion, to produce a family of $p$ individual utilities on the powerset of edges. Next, this family is aggregated, generally with the Pareto dominance, into a global preference, noted in this case $\succcurlyeq_{\Sigma P}$, on the sets of edges.

**Example 5.** By running an algorithm solving the ST/$p$Σ$u$>PARETO/MAX problem on the instance $(G, (2, u))$ described in the Example 4, we obtain the maximal set $M(S_{ST}(G), \succcurlyeq_{\Sigma P}) = \{abh, acd, ach, bcd, bch\}$, which is strictly included in $M(S_{ST}(G), \succcurlyeq_{PK})$.

The following theorem describes the relationship between the classical hierarchical aggregation $p$Σ$u$>PARETO and ours PARETO>TOSORT-VSMAX:

**Theorem 2.** *Given an undirected graph $G = (V, E)$, and a couple $(p, u)$ made up a positive number $p$ and a multi-attribute utility function $u$ from $E \times \{1, ..., p\}$ to ℝ; then every maximal solution for ST/$p$Σ$u$>PARETO/MAX is also a maximal solution for ST/PARETO>TOSORT-VSMAX/MAX. Formally:*

$$\forall x \in S_{ST}(G), x \in M(S_{ST}(G), \succcurlyeq_{\Sigma P}) \Rightarrow x \in M(S_{ST}(G), \succcurlyeq_{PK}) \quad (1)$$

Before showing this theorem, here is a lemma which describes a property of the relation $\succcurlyeq_K$:

**Lemma 1.** Given a couple $(G = (V, E), \succcurlyeq_E)$ and an element $x \in \mathcal{P}(E)$, then the relation $\succcurlyeq_K$ is transitive and:
$\exists y \in \mathcal{P}(E)$ such that $x \succcurlyeq_K y \Leftrightarrow x$ is optimal in $(\mathcal{P}(E), \succcurlyeq_K)$
$\Leftrightarrow x$ is maximal in $(\mathcal{P}(E), \succcurlyeq_K)$
Moreover, if $x \in S_{ST}(G)$, then:
$\exists y \in S_{ST}(G)$ such that $x \succcurlyeq_K y \Leftrightarrow x$ is optimal in $(S_{ST}(G), \succcurlyeq_K)$
$\Leftrightarrow x$ is maximal in $(S_{ST}(G), \succcurlyeq_K)$

**Proof:** The demonstration of the optimality (first equivalence) is immediate. What about maximality (second equivalence)? If $x$ is optimal, then $x$ is maximal. Now, what about the contrary case ? If $x$ is maximal in $(\mathcal{P}(E), \succcurlyeq_K)$ then, there 2 cases:
If there exists a $z$ such that $x \succcurlyeq_K y$, then $x$ is optimal according to the first equivalence.
Otherwise (⇔ if such a $z$ does not exist), then $\forall w \in \mathcal{P}(E), x \parallel_K w$
$\Leftrightarrow \forall w \in \mathcal{P}(E), \text{not}(x \succcurlyeq_K w)$ and $\text{not}(w \succcurlyeq_K x)$. Consequently, there is no optimal element in $(\mathcal{P}(E), \succcurlyeq_K)$. This assertion is equivalent to say that for every linear extension $\{e_1, ..., e_{|E|}\}$ of $\succcurlyeq_E$ on $E$, and for every subset $z$ of $E$, there exists a $1 \leq j \leq |E|$ verifying that: $e_j \notin x$ and $(x \cap \{e_1, ..., e_{j-1}\}) \cup \{e_j\}$ is acyclic. This is possible, if and only if $(V, E)$ is a tree and $E$ is not in $\mathcal{P}(E)$. This is a contradiction.

Hence, at last, $x$ is optimal in $(\mathcal{P}(E), \succcurlyeq_K) \Leftrightarrow x$ is maximal in $(\mathcal{P}(E), \succcurlyeq_K)$.

The transitivity of $\succcurlyeq_K$ is a direct consequence of the first equivalence. Next, both the last equivalences are true because $(\mathcal{P}(E), \succcurlyeq_K)$ verifies the Arrow choice axiom [23]:
For any $\mathcal{A}, \mathcal{B} \in \mathcal{P}(E)$ and $\mathcal{A} \subseteq \mathcal{B}$,
If $B(\mathcal{B}, \succcurlyeq_K) \cap \mathcal{A} \neq \emptyset$ then $B(\mathcal{B}, \succcurlyeq_K) \cap \mathcal{A} = B(\mathcal{A}, \succcurlyeq_K)$
(every restriction of $\mathcal{P}(E)$ conserves the optimality). ❑

**Proof (Theorem 2):** First of all, both the following assertions are false:
(a) $\forall x, y \in S_{ST}(G), x \succcurlyeq_{PK} y \Rightarrow x \succcurlyeq_{\Sigma P} y$
(b) $\forall x, y \in S_{ST}(G), x \succcurlyeq_{\Sigma P} y \Rightarrow x \succcurlyeq_{PK} y$

Indeed, for the assertion (a), it is sufficient to take the undirected graph of Figure 3, with the bicriteria utility function of Table 1.

The assertion (b) is false because PK only carries out the dichotomy between the maximal set and its complementary. So, the preferences between two non maximal elements are unknown.

We prove now the formulae (1). So, we reason by contradiction: Suppose there exists an $x \in S_{ST}(G)$ maximal for $\succcurlyeq_{\Sigma P}$, but not for $\succcurlyeq_{PK}$. This proposition is equivalent with the following one, according to Lemma 1:
$\exists x \in S_{ST}(G)$ such that: $[\forall y \in S_{ST}(G), \text{not}(y \succ_{\Sigma P} x)]$ and
$[\forall y \in S_{ST}(G), \text{not}(x \succcurlyeq_{PK} y)]$
By definition, $\text{not}(x \succcurlyeq_{PK} y) \Leftrightarrow \exists e_1 \in E \setminus x$, and $\exists e_2 \in L(x \cup \{e_1\})$ verifying $e_1 \succ_{EP} e_2$.
Now, if we take the spanning tree $y$ defined as follow: $y = x \cup \{e_1\} \setminus \{e_2\}$, then we have, because of the definition of $e_1 \succ_{EP} e_2$:

$$\forall 1 \leq i \leq p, \sum_{e \in x \setminus \{e_2\}} u(e,i) + u(e_1,i) \leq \sum_{e \in x \setminus \{e_2\}} u(e,i) + u(e_2,i), \text{ and}$$

$$\exists 1 \leq k \leq p, \sum_{e \in x \setminus \{e_2\}} u(e,k) + u(e_1,k) < \sum_{e \in x \setminus \{e_2\}} u(e,k) + u(e_2,k).$$

$\Leftrightarrow y \succ_{\Sigma P} x$. This contradicts the maximality of $x$ in $(S_{ST}(G), \succcurlyeq_{\Sigma P})$.
Hence the result. ❑

In the following, we propose an algorithm solving: GPC(ST/TOSORT-VSMAX/MAX), the global preferential consistency problem associated with ST/TOSORT-VSMAX/MAX.

## 5 GLOBAL PREFERENTIAL CONSISTENCY AND TOSORT-VSMAX

Instead of either listing all the maximal spanning trees, or finding such one tree, we will point out the removing of edges belonging to no maximal spanning tree. Especially here, we are interested in GPC(ST/TOSORT-VSMAX/MAX). Here is its definition:

GPC(ST/TOSORT-VSMAX/MAX): Let $G = (V, E)$ be an undirected graph and $\succcurlyeq_E$ be a P-acyclic preference relation on $E$ representing a TOSORT-VSMAX preference relation $\succcurlyeq$ on $\mathcal{P}(E)$. Return all the edges of $E$ belonging to a maximal spanning tree for $\succcurlyeq$, if such edges exist. Otherwise return 'no'.

Denote $S_{GPC(ST/TV/MAX)}(G, \succcurlyeq_E) \subseteq E$, the edges set outputted by an algorithm solving this problem. Then, by definition, we have the following equality:

$$S_{GPC(ST/TV/MAX)}(G, \succcurlyeq_E) = \bigcup_{x \in S_{ST/TV/MAX}(G, \succcurlyeq_E)} x. \quad (2)$$

This equality is equivalent to the conjunction of the following assertions:
(a) for all $e \in S_{GPC(ST/TV/MAX)}(G, \succcurlyeq_E) \subseteq E$, there exists $x \in S_{ST/TV/MAX}(G, \succcurlyeq_E) \subseteq \mathcal{P}(E)$, such that: $e \in x$.
(b) for all $x \in S_{ST/TV/MAX}(G, \succcurlyeq_E) \subseteq \mathcal{P}(E)$, $x \subseteq S_{GPC(ST/TV/MAX)}(G, \succcurlyeq_E)$.

The Figure 6 presents an algorithm solving this preferential consistency problem.

---

GPCORDINALSTMAX1($G = (V, E)$: undirected graph, $\succcurlyeq_E$:
P-acyclic preference relation on $E$):
return {edges set, no}
**begin**
(1) **if** ( NBCONNECTEDCOMPONENTS($G$) > 1 ) **then return** no **end if**
(2) $A \subseteq E \leftarrow \emptyset$
(3) $B \subseteq E \leftarrow E$
(4) $C(e) \subseteq E \leftarrow \emptyset$, for every $e \in E$
(5) **while** ( $B \neq \emptyset$ ) **do**
   % loop invariants: $A \cap B = \emptyset$ and $B \cap C(e) = \emptyset$
(6)   $e \leftarrow \text{CHOOSE}(M(B, \succcurlyeq_E))$
(7)   $B \leftarrow B \setminus \{e\}$
(8)   $C(e) \leftarrow \left( \bigcup_{e' \succ_E e} C(e') \cup \{e'\} \right)$
(9)   **if** ( NBCONNECTEDCOMPONENTS($V, C(e) \cup \{e\}$) <
         NBCONNECTEDCOMPONENTS($V, C(e)$) ) **then**
         $A \leftarrow A \cup \{e\}$
      **end if**
(10) **end while**
(11) **return** $A$
**end** GPCORDINALSTMAX1

---

**Figure 6.** An algorithm solving the GPC(ST/TOSORT-VSMAX/MAX) problem.

This algorithm supposes we know:
- Another algorithm NBCONNECTEDCOMPONENTS solving the counting problem of the connected components in an undirected graph. This problem is known solvable in a linear time (by a depth first search algorithm) for any given undirected graph (see e.g. [22, § 6.3 p. 90]).
- A choice strategy CHOOSE outputting an element of the input explicit set in the non-empty case. Otherwise, return 'no'.

**Example 6.** By running GPCORDINALSTMAX1 on the instance given in Figure 1 and Figure 3, we obtain as result the respective edges sets $\{a, b\}$ and $E \setminus \{b\}$.

We denote $|(G, \succcurlyeq_E)|$ the size of the instance $(G, \succcurlyeq_E)$ of ST/TOSORT-VSMAX/MAX. This size can be formulated in terms of the vertices set cardinality $m = |V|$ of graph $G$, the number of edges $n = |E|$ in $G$, and the number of arcs $p = |\succcurlyeq_E|$ in $(E, \succcurlyeq_E)$: Hence, $|(G, \succcurlyeq_E)|$ is in $O(m + n + p)$. Now, we remark that $0 \leq n \leq m^2$ and $0 \leq p \leq n^2$. Hence, $|(G, \succcurlyeq_E)|$ is in $O(m^4)$. We have the following results:

**Property 2.** The algorithm GPCORDINALSTMAX1 has a worst case time complexity, which is linear in the size of the input $(G, \succcurlyeq_E)$.

**Proof:** It is simply sufficient to see that an order of magnitude for the worst case time complexity of this algorithm GPCORDINALSTMAX1 only depends on the second loop (lines 5 to 10). The algorithm

NBCONNECTEDCOMPONENTS, solving the counting connected components problem in time linear in the size of its instance (a partial graph of $G$), is then in $O(m + n)$. Consequently, the worst case time complexity of the conditional instruction '**if …end if**' (line 9) is about $m + n$. It is similar for:
- line 6, where the choice strategy necessitates a greedy search of maximal edge, solvable in the worst case in $O(n)$
- line 8, where the maximum number of possible unions is about cardinality of $E$, i.e. in $O(n)$.

At last, the body of the $2^{nd}$ loop runs in the worst case in $O(m + n)$ times. Now, the number of loops is equal to the number of edges; and proves that the complexity of the algorithm GPCORDINALSTMAX1 is in $O((m + n) \cdot n) \approx O(m^4)$, i.e. linear in the input size $|(G, \succcurlyeq_E)|$. ❑

**Theorem 3** *The algorithm GPCORDINALSTMAX1 returns the whole MAX-consistent edges (and only them) for maximal spanning trees of the ST/TOSORT-VSMAX/MAX problem, from an instance $((V, E), \succcurlyeq_E)$, if such trees exist. Otherwise returns 'no'.*

The logic underlying this algorithm consists in putting an edge $e \in E$ in a best scenario of choice, in order to elaborate a linear extension of $\succcurlyeq_E$ (= the minimal number assigned to $e$ among the linear extensions). Such a best scenario consists in choosing $e$ as soon as possible, during the topological sort. For that, the topological sorting algorithm has to number every better edge $e'$ than $e$ for $\succcurlyeq_E$; next the edge $e'$ is numbered iff every better edge than $e'$ is numbered, and so on. In the best case, when $e$ is numbered, if the number of connected components decreases when we add $e$ to the already numbered edges, then $e$ can be chosen to belong to a maximal spanning tree for a TOSORT-VSMAX preference relation. Indeed, this best scenario may then be completed in a maximal spanning tree, by iteratively choosing any maximal remaining edge.

Before showing this theorem, here is a lemma which will help us in the demonstration.

**Lemma 2.** Given an instance $(G, \succcurlyeq_E)$, with $\succcurlyeq_E$ P-acyclic, denote $C(e)$ the edges of $E$ for which there exists a path of strict preferences towards $e$: $C(e) = \{f \in E$ such that: $\exists f^{(1)}, …, f^{(p)} \in E$, with $p \geq 0$, and $f \succ_E f^{(1)} \succ_E … \succ_E f^{(p)} \succ_E e\}$. Then, for every $A \subseteq E$, $C(e) \setminus A \neq \varnothing \Rightarrow M(C(e) \setminus A, \succcurlyeq_E) \subseteq M(E \setminus A, \succcurlyeq_E)$

**Proof:** First of all, let us clarify the set $E \setminus C(e)$:
$E \setminus C(e) = \{f \in E$ such that: There exists <u>no</u> path of strict preferences from $f$ to $e$ in $(E, \succcurlyeq_E)\}$
$= \{f \in E$ such that: There is no path of strict preferences from $f$ to $e_1$ in $(E, \succcurlyeq_E), \forall e_1 \in C(e)\}$.
Indeed, if such a path existed from $f$ to $e_1$, and – by definition of $C(e)$ – from $e_1$ to $e$, then there would exist a path of strict preferences from $f$ to $e$.

Show now lemma: Suppose that $C(e) \setminus A \neq \varnothing$. Then every edge $f \in M(C(e) \setminus A, \succcurlyeq_E)$ verifies: $\forall f_1 \in C(e) \setminus A$, not$(f_1 \succ_E f)$
$\Rightarrow$ There exists <u>no</u> path of strict preferences from $f_1$ to $f$ in $(C(e) \setminus A, \succcurlyeq_E)$.

Hence, <u>if</u> some edges are added to $C(e) \setminus A$ – in this case $E \setminus (C(e) \cup A)$ – for which there exists <u>no</u> path of strict preferences from $f_2 \in E \setminus (C(e) \cup A)$ to $e_1 \in C(e) \setminus A$, <u>then</u> it won't also exist a path from $f_2$ to $f$. At last, $\forall f \in M(C(e) \setminus A, \succcurlyeq_E), \forall f_1 \in E \setminus A$, not$(f_1 \succ_E f)$. This shows lemma. ❑

**Proof (Theorem 3):** Firstly, we have the following equivalence, because of the Properties 1 (*b*) translated by the first line of the algorithm: GPCORDINALSTMAX1$(G, \succcurlyeq_E)$ = 'no' $\Leftrightarrow S_{ST}(G) = \varnothing$.

Then point out on the first part of the proposition: Suppose GPCORDINALSTMAX1$(G, \succcurlyeq_E) \neq$ 'no', and show – by help of Lemma 2 – formulae (2):

$$\text{GPCORDINALSTMAX1}(G, \succcurlyeq_E) = \bigcup_{x \in S_{ST/TV/MAX}(G, \succcurlyeq_E)} x .$$

Direct inclusion: For any $e \in$ GPCORDINALSTMAX1$(G, \succcurlyeq_E) \subseteq E$, there exists $x \in S_{ST/TV/MAX}(G, \succcurlyeq_E) \subseteq \mathcal{P}(E)$, such that: $e \in x$. Indeed, such an $x$ can be designed by using the strategy described in the previous remark with a topological sort of $(E, \succcurlyeq_E)$. So, as long as we are not at an iteration $k$ such that $e$ is maximal in the set $B_k$ of not yet numbered edges, then, during iterations $i < k$, the choice strategy consists in taking as current edge $e_i$ a maximal edge for $(C(e) \setminus (E \setminus B_i), \succcurlyeq_E)$, with $C(e) \setminus (E \setminus B_i) \subseteq B_i$. According to the above lemma, $M(C(e) \setminus (E \setminus B_i), \succcurlyeq_E) \subseteq M(B_i, \succcurlyeq_E)$. Therefore, this strategy is available, and the iteration $k = |C(e)|$. During the iteration $k$, given $e$ decreases the number of connected components in $C(e)$ – because $e$ is in GPCORDINALSTMAX1$(G, \succcurlyeq_E)$ and then verifies the condition of line 9 in GPCORDINALSTMAX1 –, then $e$ is chosen to be added to $A_{k-1}$, the current tree. Next, during iterations $i > k$, the topological sort algorithm takes as current edge, any edge of $M(B_i, \succcurlyeq_E)$. At last, the elaborated linear extension can be associated to a utility function (see sketch of proof of Theorem 1) and next used as instance of an algorithm solving ST/$\Sigma u$/OPT, which necessarily returns a solution $x \neq$ 'non', containing $e$ and then maximal for $(G, \succcurlyeq_E)$.

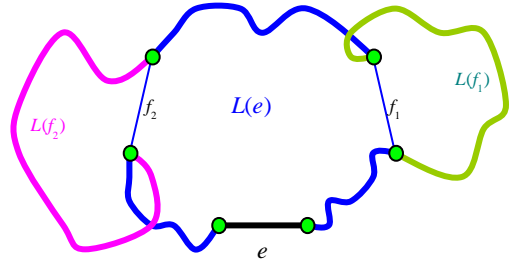

Legend:
Blue : $L(e)$ is an undirected path in $C(e) \subseteq E$ between both the ends of $e$.
$f_1$ and $f_2$ are two edges of $L(e) \setminus x$
$L(f_1)$ and $L(f_2)$ are 2 undirected paths in $x$ between both the ends of respectively $f_1$ and $f_2$.
The edges of the bold path is included in $x$.

**Figure 7.** Illustration for demonstration of Theorem 3.

Converse inclusion: For every $x \in S_{ST/TV/MAX}(G, \succcurlyeq_E) \subseteq \mathcal{P}(E)$, $x \subseteq S_{GPC(ST/TV/MAX)}(G, \succcurlyeq_E)$. Indeed, reason by contradiction. Suppose that:

$\exists x \in S_{ST/TV/MAX}(G, \succcurlyeq_E)$ and $x \not\subset S_{GPC(ST/TV/MAX)}(G, \succcurlyeq_E)$.
$\Leftrightarrow \exists e \notin S_{GPC(ST/TV/MAX)}(G, \succcurlyeq_E)$ although: $e \in x$, and $x \in S_{ST/TV/MAX}(G, \succcurlyeq_E)$.
$\Leftrightarrow$ The number of connected components does not decrease if we add $e$ in $C(e)$, according to line 9 of GPCORDINALSTMAX1
$\Leftrightarrow$ There exists in $C(e)$ an undirected path $L(e)$ between both the ends of $e$.

So that $e$ should be chosen during the design of $x$, because $e \in x$, it is necessary that $e$ be maximal at an iteration $k \leq |E|$, if we use the Kruskal's algorithm to solve ST/TO-SORT-VSMAX/MAX. At this iteration, $e \in A_k$, the tree at iteration $k$, and $e$ decreases the number of connected components in $A_{k-1}$. If $e \in M(E, \succeq_E)$, then $e \in S_{GPC(ST/TV/MAX)}(G, \succeq_E)$, this is a contradiction with the initial assumption. Accordingly, $C(e) \neq \varnothing$. Moreover, every edge of $C(e)$ has already been chosen in the scenario of the topological sort algorithm during iterations $i < k$, in order that $e$ be maximal during the iteration $k$. It is sure that $C(e) \nsubseteq x$ because it would exist an undirected path $L(e) \cup \{e\}$ in $x$; that is contradictory with $x$ is a tree. Hence $C(e) \setminus x \neq \varnothing$. And for every $f \in C(e) \setminus x$, $f$ has not been added to $x$ because during the iteration $i < k$ where it has been chosen, there already exists an undirected path $L(f)$ in $A_i \subseteq x$ between the ends of $f$.

Now, the edge set $(C(e) \cap x) \cup \left( \bigcup_{f \in C(e) \setminus x} L(f) \right)$ is an undirected path (or contains such a path) in $x$ between the ends of $e$, making up with $e$ an undirected path in $x$ (see Figure 7). This contradicts the assumption $x$ is a tree. That demonstrates the converse inclusion. ❑

## 6 CONCLUSION AND PERSPECTIVES

One of the limits, devolved upon decision processes based on listing of preferred solutions suggested by Perny & Spanjaard [18] to solve ordinal combinatorial problems, was the intractability of large size inputs. We introduced another kind of computable problems, preferential consistency ones. Their outputs can be processed in real-time by a human being (i.e. linear in the input size). These computable problems are based both on the notion of consistency pointing out by constraint programming (CP), and on the notion of choice investigated in decision aiding (DA). In the case of maximal spanning trees problems satisfying the TOSORT-VSMAX condition, we proposed an algorithm solving the global consistency problem, with a linear worst case time complexity in its input size.

One of the aims of this article is to bring together the CP and OR-DA communities, to process more efficiently combinatorial problems exploiting complex preferences. Their mutual contributions open a new way of interactive solving of semi-structured combinatorial problems. Consequently, the perspectives are numerous:

At first, with preferential consistency: Global preferential consistency can be used in an interactive decision process, where the user makes some local decisions (choice), and where the DSS is restricted to remove preferential inconsistent domain-values. However, such support systems may not always warrant a preferred solution for the initial instance. Consequently, we have explored this way, for example by identifying domain-values which are in all preferred solutions or, by investigating rational choice theory [23] to identify some sufficient properties so that the decision process always returns a preferred solution for the initial instance, if such a solution exists.

Next, with efficient spanning trees problems and the particular compact representation of preferences used in this article: We have been scrutinizing the concept of *expressive power* of a compact representation. Any kind of compact representations models only a subset of preference relations. For example, utility functions model only total preorders. In order to better understand the type of compact representation used in this article, we focus our researches on its expressive power for spanning trees problems.

At last, with applications: What makes a good theory, it is its applicability to real world problems. The possible applications are numerous. And at this time, we work on an autonomous electrical network designing problem allowing several – not necessary cardinal – criteria. Shortly, these problems arise in isolated regions as some Pacific islands or in remote villages in rainforest. The isolation of these populations implies that the continuous supply of fossil fuels is very expensive to the community, and exorbitantly expensive if you wanted to connect to an existing electricity grid. Renewable energies form a more interesting both in terms of costs (a barrel of oil more and more expensive, and means of delivery prohibitive as boat (sometimes pirogue), helicopter or plane), in terms of noise and soil pollution, etc. These problems necessitate very complex preferential information as inhabitants opinions, cost, environmental and aesthetic criteria.